\documentclass{bmvc2k}

\title{Revisiting Temporal Modeling for Video-based Person ReID}

\addauthor{Jiyang Gao}{jiyangga@usc.edu}{1}
\addauthor{Ram Nevatia}{nevatia@usc.edu}{1}

\addinstitution{
 University of Southern California\\
 Los Angeles, CA, USA
}
\addinstitution{
 University of Southern California\\
 Los Angeles, CA, USA
}
\runninghead{J. Gao}{Revisiting Temporal Modeling for Video-based Person ReID}

\def\eg{\emph{e.g}\bmvaOneDot}

\def\etal{\emph{et al}\bmvaOneDot}

\begin{document}

\maketitle

\begin{abstract}
Video-based person reID is an important task, which has received much attention in recent years due to the increasing demand in surveillance and camera networks. A typical video-based person reID system consists of three parts: an image-level feature extractor (\eg CNN), a temporal modeling method to aggregate temporal features and a loss function. Although many methods on temporal modeling have been proposed, it is hard to directly compare these methods, because the choice of feature extractor and loss function also have a large impact on the final performance. We comprehensively study and compare four different temporal modeling methods (temporal pooling, temporal attention, RNN and 3D convnets) for video-based person reID. We also propose a new attention generation network which adopts temporal convolution to extract temporal information among frames. The evaluation is done on the MARS dataset, and our methods outperform state-of-the-art methods by a large margin. Our source codes are released at \color{magenta}{\url{https://github.com/jiyanggao/Video-Person-ReID}}. 
\end{abstract}

\section{Introduction}
Person re-Identification (re-ID) tackles the problem of retrieving a specific person (\emph{i.e.} query) in different images or videos, possibly taken from different cameras in different environments. It has received increased attentions in recent years due to the increasing demand of public safety and rapidly growing surveillance camera networks. Specifically, we focus on video-based person re-ID, that is, given a query video of one person, the system tries to identify this person in a set of gallery videos. 

Most of the recent existing video-based person reID methods are based on deep neural networks \cite{zhou2017see,liu2017quality,mclaughlin2016recurrent}. Typically, three important parts have large impacts on a video-based person reID system: an image-level feature extractor (typically a Convolutional Neural Network, CNN), a temporal modeling module to aggregate image-level features and a loss function to train the network. During test, the query video and gallery videos are encoded to feature vectors using the aforementioned system, and then differences between them (usually L2 distances) are calculated to retrieve the top-N results. Recent work \cite{liu2017quality,zhou2017see,mclaughlin2016recurrent} on video-based person reID mostly focuses on the temporal modeling part, \emph{i.e.}, how to aggregate a sequence of image-level features into a clip-level feature. 

Previous work on temporal modeling methods on video-based person reID falls into two categories: Recurrent Neural Network (RNN) based and temporal attention based. In RNN-based methods, McLanghlin \etal \cite{mclaughlin2016recurrent} proposed to use an RNN to model the temporal information between frames; Yan \etal \cite{yan2016person} also used an RNN to encode sequence features, where the final hidden state is used as video representation. In temporal attention based methods, Liu \etal \cite{liu2017quality} designed a Quality Aware Network (QAN), which is actually an attention weighted average, to aggregate temporal features; Zhou \etal \cite{zhou2017see} proposed to encode the video with temporal RNN and attention. Besides, Hermans \etal \cite{hermans2017defense} adopted a triplet loss function and a simple temporal pooling method, and achieved state-of-the-art performance on the MARS \cite{zheng2016mars} dataset. 

Although extensive experiments have been reported on the aforementioned methods, it's hard to directly compare the influence of temporal modeling methods, as they used different image level feature extractors and different loss functions, these variations can affect the performance significantly. For example, \cite{mclaughlin2016recurrent} adopted a 3-layer CNN to encode the images; \cite{yan2016person} used hand-crafted features; QAN \cite{liu2017quality} extracted VGG \cite{simonyan2014very} features as image representations.  

In this paper, we explore the effectiveness of different temporal modeling methods on video-based person re-ID by fixing the image-level feature extractor (ResNet-50 \cite{he2016deep}) and the loss function (triplet loss and softmax cross-entropy loss) to be the same. Specifically, we test four commonly used temporal modeling architectures: temporal pooling, temporal attention \cite{liu2017quality,zhou2017see}, Recurrent Neural Network (RNN) \cite{mclaughlin2016recurrent,yan2016person} and 3D convolution neural networks \cite{hara3dcnns}. A 3D convolution neural network \cite{hara3dcnns} directly encodes an image sequence as a feature vector; we keep the network depth the same as for the 2D CNN for fair comparison. We also propose a new attention generation network which adopts temporal convolution to extract temporal information. We perform experiments on the MARS \cite{zheng2016mars} dataset, which is the largest video-based person reID dataset available to date. The experimental results show that our method outperforms state-of-the-art models by a large margin.

In summary, our contributions are two-fold: First, we comprehensively study four commonly used temporal modeling (temporal pooling, temporal attention, RNN and 3D conv) methods for video-based person reID on MARS. We will release the source codes. Second, we propose a novel temporal-conv based attention generation network, which achieves the best performance among all temporal modeling methods; with the help of strong feature extractor and effective loss functions, our system outperforms state-of-the-art methods by a large margin. 

In the following, we first discuss related work in Sec \ref{sec:work}, then present the overall person reID system architecture in Sec \ref{sec:method}, and describe the temporal modeling methods in detail. In Sec \ref{sec:eval}, we show the experiments and discuss the results. 

\section{Related Work}
\label{sec:work}
In this section, we discuss related work, including video-based and image-based person reid and video temporal analysis.

\textbf{Video-based person reID.} Previous work on temporal modeling methods on video-based person reID fall into two categories: Recurrent Neural Network (RNN) based and temporal attention based. McLanghlin \etal \cite{mclaughlin2016recurrent} first proposed to model the temporal information between frames by RNN, the average of RNN cell outputs is used as the clip level representation. Similar to \cite{mclaughlin2016recurrent}, Yan \etal \cite{yan2016person}also used RNN to encode sequence features, the final hidden state is used as video representation. Liu \etal \cite{liu2017quality} designed a Quality Aware Network (QAN), which is essentially an attention weighted average, to aggregate temporal features; the attention scores are generated from frame-level feature maps. Zhou \etal \cite{zhou2017see} and Xu \etal \cite{shuangjiejointly} proposed to encode encode the video with temporal RNN and attention. Chung \etal \cite{chung2017two} presented a two-stream network which models both RGB images and optical flows, simple temporal pooling is used to aggregate the features. Recently, Zheng \etal \cite{zheng2016mars} built a new dataset MARS for video-based person reID, which becomes the standard benchmark on this task. 

\textbf{Image-based person reID.} Recent work on image-based person reID improves the performance by mainly two directions: image spatial modeling and loss function for metric learning. In the direction of spatial feature modeling, Su \etal \cite{su2017pose} and Zhao \etal \cite{zhao2017spindle} used human joints to parse the image and fuse the spatial features. Zhao \etal \cite{zhao2017deeply} proposed a part-aligned representation for handling the body part misalignment problem. As for loss functions, typically, hinge loss in a Siamese network and identity softmax cross-entropy loss function are used. To learn an effective metric embedding, Hermans \etal \cite{hermans2017defense} proposed a modified triplet loss function, which selects the hardest positive and negative for each anchor sample, and they achieved state-of-the-art performance.

\textbf{Video temporal analysis.} Besides person reID work, temporal modeling methods in other fields, such as video classification \cite{Karpathy_2014_CVPR}, temporal action detection \cite{Shou_2016_CVPR, gao2017red}, are also related. Karpathy \etal \cite{Karpathy_2014_CVPR} designed a CNN network to extract frame-level features and used temporal pooling method to aggregate features. Tran \etal \cite{tran2015learning} proposed a 3D CNN network to extract spatio-temporal features from video clips. Hara \etal \cite{hara3dcnns} explored the ResNet \cite{he2016deep} architecture with 3D convolution. Gao \etal \cite{gao2017turn, Gao_2017_cbr} proposed a temporal boundary regression network to localize actions in long videos.

\begin{figure*}[h]
  \centering
    \includegraphics[width=0.90\textwidth]{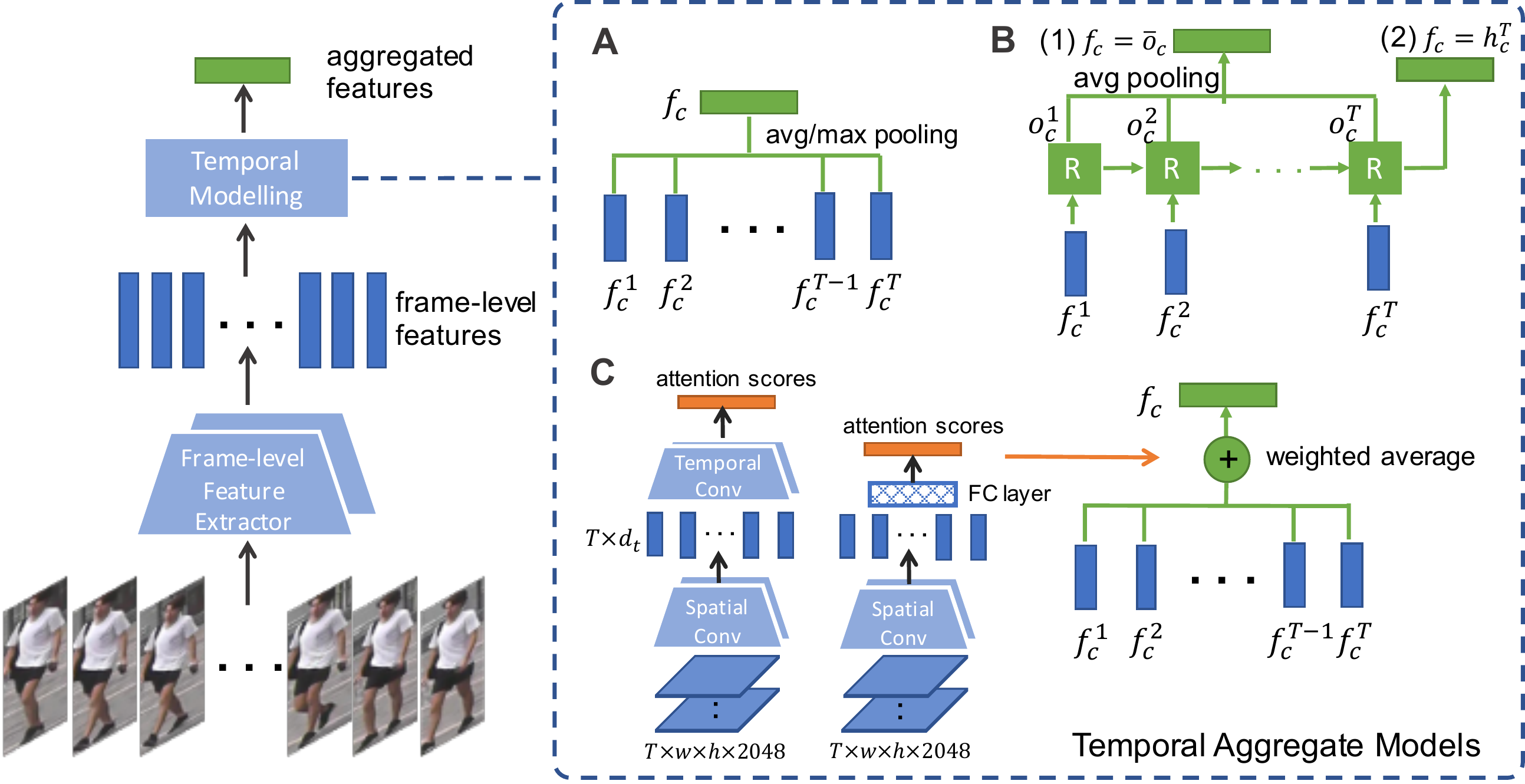}
    \caption{Three temporal modeling architectures (A: temporal pooling, B: RNN and C: temporal attention) based on an image-level feature extractor (typically a 2D CNN). For RNN, final hidden state or average of cell outputs is used as the clip-level representation; For temporal attention, two types of attention generation network are shown: ``spatial conv + FC \cite{liu2017quality}" and ``spatial conv + temporal conv". }
      \label{fig:model}
\end{figure*}

\section{Methods}
\label{sec:method}
In this section, we introduce the overall system pipeline and detailed configurations of the temporal modeling methods. The whole system could be divided into two parts: a video encoder which extract visual representations from video clips, and a loss function to optimize the video encoder and a method to match the query video with the gallery videos. A video is first cut into consecutive non-overlap clips $\{c_k\}$, each clip contains $T$ frames. The clip encoder takes the clips as inputs and outputs a D-dimensional feature vector $f_c$ for each clip. The video level feature is the average of all clip level features.

\subsection{Video Clip Encoder}
To build a video clip encoder, we consider two types of convolutional neural network (CNN): (1) 3D CNN and (2) 2D CNN with temporal aggregation method. 3D CNN directly takes a video clip $c$ which contains $n$ frames as input and output a feature vector $f_c$, while 2D CNN first extracts a sequence of image-level features $\{f_c^t\}, t\in[1,n]$, and then $\{f_c^t\}$ are aggregated into a single vector $f_c$ by a temporal modeling method. 

\textbf{3D CNN.} For 3D CNN, we adopt 3D ResNet \cite{hara3dcnns} model, which adopts 3D convolution kernels with ResNet architecture \cite{he2016deep} and is designed for action classification. We replace the final classification layer with person identity outputs and use the pre-trained parameters (on Kinetics \cite{kay2017kinetics}). The model takes $T$ consecutive frames (\emph{i.e.} a video clip) as input, and the layer before final classification layer is used as the representation of the person. 

For 2D CNN, we adopt a standard ResNet-50 \cite{he2016deep} model as an image level feature extractor. Given an image sequence (\emph{i.e.} a video clip), we input each image into the feature extractor and output a sequence of image level features $\{f_c^t\}, t\in[1,T]$, which is a $T \times D$ matrix, n is clip sequence length, $D$ is image level feature dimension.  Then we apply a temporal aggregation method to aggregate the features into a single clip level feature $f_c$, which is a D-dimensional vector. Specifically, we test three different temporal modeling methods: (1) temporal pooling, (2) temporal attention, (3) RNN; the architectures of these methods are shown in Figure \ref{fig:model}. 

\textbf{Temporal pooling.} In temporal pooling model, we consider max pooling and average pooling. For max pooling, $f_c=max_t(f_c^t)$; for average pooling, $f_c=\frac{1}{T} \sum_{t=1}^n f_c^t$. 

\textbf{Temporal attention.} In temporal attention model, we apply an attention weighted average on the sequence of image features. Given that the attention for clip $c$ is $a_c^t, t\in[1,T]$, then 
\begin{equation}
   f_c = \frac{1}{T} \sum_{t=1}^n a_c^t f_c^t
\end{equation}
The tensor size of last convolutional layer in Resnet-50 is $[w,h,2048]$, w and h depend on input image size. The attention generation network takes a sequence of image-level features $[T,w,h,2048]$ as inputs, and outputs $T$ attention scores. We design two types of attention networks. (1) ``spatial conv + FC \cite{liu2017quality}": we apply a spatial conv layer (kernel width = w, kernel height = h, input channle number =$2048$, output channel number = $d_t$, short for $\{w,h,2048,d_t\}$) and a Fully-Connected (FC) (input channel = $d_t$, output channel = 1) layer on the aforementioned output tensor; the output of the conv layer is a scalar vector $s_c^t, t\in [1,T]$, which is used as the score for the frame $t$ of clip $c$. (2) ``spatial + temporal conv": first a conv layer with shape $\{w,h,2048,d_t\}$ is applied, then we get a $d_t$-dimensional feature for each frame of the clip, we apply a temporal conv layer $\{3, d_t, 1\}$ on these frame-level features to generate temporal attentions $s_c^t$. The two networks are shown in Figure \ref{fig:model} (C). 

Once we have $s_c^t$, there are two ways of calculating the final attention scores $a_c^t$: (1) Softmax function \cite{zhou2017see}, 
\begin{equation}
    a_c^t=\frac{e^{s_c^t}}{\sum_{i=1}^{T}e^{s_c^i}} 
\end{equation}
and (2) a Sigmoid function + L1 normalization \cite{liu2017quality}, 
\begin{equation}
  a_c^t=\frac{\sigma(s_c^t)}{\sum_{i=1}^{T}\sigma(s_c^i)}  
\end{equation}
where $\sigma$ means the Sigmoid function.

\textbf{RNN.} An RNN cell encodes one image feature in a sequence at one time step $t$ and passes the hidden state $h_t$ into the next time step. We consider two methods of aggregating a sequence of image features into a single clip feature $f_c$. The first method directly takes the hidden state $h^{T}$ at the last time step, $f_c=h_c^{T}$, as shown in Figure \ref{fig:model} (B). The second method calculates the average of the RNN outputs $\{o^t\}, t\in[1,n]$, that is $f_c=\frac{1}{T}\sum_{t=1}^{T} o_c^t$. We test two types of RNN cell: Long Short-Term Memory (LSTM) and Gated Recurrent Unit (GRU). Other settings can be found in Section \ref{sec:evalsetup}.

\subsection{Loss Functions}
We use a triplet loss function and a Softmax cross-entropy loss function to train the networks. The triplet loss function we use was originally proposed in \cite{hermans2017defense}, and named as Batch Hard triplet loss function. To form a batch, we randomly sample P identities and randomly sample K clips for each identity (each clip contains $T$ frames); totally there are $PK$ clips in a batch. For each sample $a$ in the batch, the hardest positive and the hardest negative samples within the batch are selected when forming the triplets for computing the loss $L_{triplet}$. 
    \begin{align}\label{eq:loss_bh}
        L_{triplet} = \overbrace{\sum\limits_{i=1}^{P} \sum\limits_{a=1}^{K}}^{\textnormal{all anchors}}
            \Big[m & + \hspace*{-5pt} \overbrace{\max\limits_{p=1 \dots K} \hspace*{-5pt} D\left(f_{i,a}), f^i_p)\right)}^{\textnormal{hardest positive}} \\
                   & - \hspace*{-5pt} \underbrace{\min\limits_{\substack{j=1 \dots P \\ n=1 \dots K \\ j \neq i}} \hspace*{-5pt} D\left(f_{i,a}), f_{j,n})\right)}_{\textnormal{hardest negative}} \Big]_+,\nonumber
    \end{align}%
The Softmax cross-entropy loss function $L_{softmax}$ encourages the network to classify the $PK$ clips to the correct identities. 
\begin{equation}
    L_{softmax} = -\frac{1}{PK}\sum_{i=1}^{P}\sum_{a=1}^{K}p_{i,a}\mbox{log}q_{i,a}
\end{equation}
where $p_{i,a}$ and $q_{i,a}$ are the groundtruth identity and prediction of sample $\{i,a\}$. The total loss $L$ is the combination of these two losses. 
\begin{equation}
L=L_{softmax}+L_{triplet}
\end{equation}

\subsection{Similarity Calculation for Testing}
\label{sec:sim}
As mentioned before, a video is cut into consecutive non-overlapping clips $\{c_k\}$, each clip containing $T$ frames. During testing, we extract clip level representation for each clip in a video, the video level representation is the average of all clip level representations. L2 distance is used to measure the similarity among videos.

\section{Evaluation}
\label{sec:eval}
In this section, we list the evaluation settings and discuss experimental results.

\subsection{Evaluation Settings}
\label{sec:evalsetup}
We introduce evaluation metric, dataset and image baseline models. Implementation details are also shown.

\textbf{Metric.} We use the standard evaluation metrics: mean average precision score (mAP) and the
cumulative matching curve (CMC) at rank-1, rank-5, rank-10 and rank-20.

\textbf{Dataset.} We test all models on the MARS dataset \cite{zheng2016mars}, which is the largest video-based person reID dataset to date. MARS consist of ``tracklets" which have been grouped into person IDs. It contains 1261 IDs and around 20000 tracklets. The train and test are split evenly. 

\textbf{Implementations.} Standard ResNet-50 \cite{he2016deep} pretrained on ImageNet is used as the 2D CNN and 3D ResNet-50 \cite{hara3dcnns} pretrained on Kinectics \cite{kay2017kinetics} is used as the 3D CNN video encoder. The video frames are resized to 224 $\times$ 112. Adam \cite{kingma2014adam} is used to optimize the networks. Batch size is set to 32; if the total memory usage exceeds GPU memory limit, we accordingly reduce the batch size to the maximum possible size. In a batch, we select $P=4$ samples for each identity. We test the learning rate with 0.0001 and 0.0003 for different models to achieve the best performance. 

\textbf{Image-based baseline models.} We provide an image-based baseline model to test the effectiveness of temporal modeling. This model is similar to \cite{hermans2017defense}, but use an additional Softmax cross-entropy loss. Specifically, the sequence length of the clip is set to $T=1$, no temporal modeling method is used. The same loss function (triplet loss and cross-entropy loss) and ResNet-50 network is adopted. During testing, the same similarity calculation procedure is used as described in Sec \ref{sec:sim}.

\subsection{Experiments on MARS}
In this part, we report the performance of 3D CNN, temporal pooling, temporal attention and RNN separately and then discuss the experimental results.

\textbf{3D CNN.} ResNet3D-50 \cite{hara3dcnns} is used as the test architecture, which has the same number of layer as ResNet-50. ResNet3D-50  is a fully convolutional network, so we can change the input image sequence length. The input image height and width are kept as $224$ and $112$. Due to limited GPU memory, we only test sequence length for $T=4$ and $T=8$. We set the learning rate as 0.0001. The results are shown in Table \ref{tbl:3d}. We can see that $T=4$ performs better $T=8$.

\begin{table}[h]\small
\centering
\caption{Comparison of different sequence length $T=4,8$ with 3D CNN.}
\label{tbl:3d}
\begin{tabular}{l|ccccc}
\hline
\multicolumn{1}{c|}{} & \multicolumn{1}{c}{mAP} & \multicolumn{1}{c}{CMC-1} & \multicolumn{1}{c}{CMC-5} & CMC-10 & CMC-20 \\ \hline
T=4           &    \textbf{70.5} &	\textbf{78.5}	& \textbf{90.9}	& \textbf{93.9}	& \textbf{95.9}   \\ \hline
T=8           &    69.1	& 78.8	& 89.8	& 93.2	& 95.1   \\ \hline
\end{tabular}
\end{table}

\textbf{Temporal pooling.} The average pooling and max pooling model are tested with different sequence lengths. The learning rate is set to 0.0003, as we found this rate achieves the best performance. First, we compare average pooling and max pooling with the same sequence length $T=8$, as shown in Table \ref{tbl:maxavg}.

\begin{table}[h]\small
\centering
\caption{Comparison between average pooling and max pooling with sequence length $T=8$.}
\label{tbl:maxavg}
\begin{tabular}{l|ccccc}
\hline
\multicolumn{1}{c|}{} & \multicolumn{1}{c}{mAP} & \multicolumn{1}{c}{CMC-1} & \multicolumn{1}{c}{CMC-5} & CMC-10 & CMC-20 \\ \hline
avg pool           & \textbf{76.2} &	\textbf{82.9} &	\textbf{93.7} &	\textbf{95.4} &	\textbf{96.8}       \\ \hline
max pool           &    74.5	& 83.1	& 93.3&	95.6&	96.7     \\ \hline
\end{tabular}
\end{table}

We can see that average pooling consistently works better than max pooling. Next, we test average pooling with different sequence length $T=1, 2, 4, 8$. $T=1$ is image-based baseline model and does need to use a temporal aggregation method. The results are shown in Table \ref{tbl:avglength}, we can see that $T=4$ achieves the best performance.

\begin{table}[h]\small
\centering
\caption{Comparison of different sequence length $T=1,2,4,8$ with average pooling.}
\label{tbl:avglength}
\begin{tabular}{l|ccccc}
\hline
\multicolumn{1}{c|}{} & \multicolumn{1}{c}{mAP} & \multicolumn{1}{c}{CMC-1} & \multicolumn{1}{c}{CMC-5} & CMC-10 & CMC-20 \\ \hline
T=1           &     74.1 &	81.3&	92.6&	94.8&	96.7 \\ \hline
T=2           & 76.1 &	83.3	&93.2	& 95.5	& 96.8   \\ \hline
T=4       & 76.5	&83.5	&93.0	&95.3	&96.8 \\\hline
T=8           &    76.2 &	82.9 &	93.7 &	95.4 &	96.8    \\ \hline
\end{tabular}
\end{table}

\textbf{Temporal attention.} We evaluate two temporal attention models mentioned in Sec \ref{sec:method}: ``spatial conv + FC" and ``spatial conv + temporal conv" and two attention generation functions (Softmax and Sigmoid). The sequence length is set to $T=4$ (according to the experiments above, $T=4$ achieves the best results), learning rates are both set to 0.0003, and $d_t$ is set to 256. The comparison between attention generation functions are shown in Table \ref{tbl:attfunc}, ``spatial conv + FC" is used. We can see that ``softmax" and ``sigmoid" performs similarly. The comparison between attention generation network is shown in Table \ref{tbl:attnet}, ``softmax" is used as attention generation function. It can be seen that ``spatial conv + temporal conv" performs better than ``spatial conv + FC", which shows the effectiveness of using temporal convolution to capture information amongst frames.

\begin{table}[h]\small
\centering
\caption{Comparison between the Softmax and Sigmoid attention generation function.}
\label{tbl:attfunc}
\begin{tabular}{l|ccccc}
\hline
\multicolumn{1}{c|}{} & \multicolumn{1}{c}{mAP} & \multicolumn{1}{c}{CMC-1} & \multicolumn{1}{c}{CMC-5} & CMC-10 & CMC-20 \\ \hline

softmax          & 75.8	& 82.7	& 92.8	& 95.4	& 97       \\ \hline
sigmoid          & 76.1	& 82.8	& 93.6	& 95.4	& 96.8       \\ \hline

\end{tabular}
\end{table}

\begin{table}[h]\small
\centering
\caption{Comparison between the ``spatial conv + FC" and ``spatial conv + temporal conv", Softmax is used for attention generation.}
\label{tbl:attnet}
\begin{tabular}{l|ccccc}
\hline
\multicolumn{1}{c|}{} & \multicolumn{1}{c}{mAP} & \multicolumn{1}{c}{CMC-1} & \multicolumn{1}{c}{CMC-5} & CMC-10 & CMC-20 \\ \hline
spatial conv + FC           & 75.8 &	82.7	&93.0&	95.1&	96.5       \\ \hline
spatial + temporal conv         & \textbf{76.7} &	\textbf{83.3}	& \textbf{93.8}	& \textbf{96.0}	& \textbf{97.4} \\ \hline
\end{tabular}
\end{table}

\textbf{RNN.} We first test the RNN model with different output selections: final hidden state $f_c=h_c^T$ and cell output average pooling $f_c=\overline{o_c^T}$, as described in Sec \ref{sec:method}. Sequence length is set to 8, and learning rate is set to 0.0001. LSTM is used as the basic RNN cell. Different hidden state size (512,1024,2048) are tested. The results for final hidden state and cell output average pooling are shown in Table \ref{tbl:final} and Table \ref{tbl:avgoutput} respectively. We can see that ``cell outputs average" works generally better than ``final hidden state". $D_h=512$ achieves the best performance in both two models.

\begin{table}[h]\small
\centering
\caption{Comparison of different hidden state size $D_h=512,1024,2048$ with using ``final hidden state" as the clip level representation.}

\label{tbl:final}
\begin{tabular}{l|ccccc}
\hline
\multicolumn{1}{c|}{} & \multicolumn{1}{c}{mAP} & \multicolumn{1}{c}{CMC-1} & \multicolumn{1}{c}{CMC-5} & CMC-10 & CMC-20 \\ \hline
$512$           &     \textbf{69.8} &	\textbf{79.3}	& \textbf{91.7}&	\textbf{93.3}&	\textbf{96.2}  \\ \hline
$1024$           &    66.7 &	76.4	&89.0	&92.2	& 94.7   \\ \hline
$2048$           &     67.0 &	77.7 &	89.6 &	92.8 &	94.9  \\ \hline
\end{tabular}
\end{table}

\begin{table}[h]\small
\centering
\caption{Comparison of different hidden state size $D_h=512,1024,2048$ with using the ``cell outputs average" as the clip level representation.}

\label{tbl:avgoutput}
\begin{tabular}{l|ccccc}
\hline
\multicolumn{1}{c|}{} & \multicolumn{1}{c}{mAP} & \multicolumn{1}{c}{CMC-1} & \multicolumn{1}{c}{CMC-5} & CMC-10 & CMC-20 \\ \hline
$512$           &     72.0 &	\textbf{80.4}	&\textbf{92.7}&	\textbf{94.9}	& \textbf{96.9}  \\ \hline
$1024$           &     \textbf{72.1}	& 81	& 92.5	& 94.8	& 96.5  \\ \hline
$2048$           &     70.1	& 79.5	& 91  &93.8	& 95.5  \\ \hline
\end{tabular}
\end{table}

To test different types of RNN cells (LSTM and GRU), we fix hidden state size to be $D_h=512$, sequence length to be $T=8$ and use ``cell outputs average" as the clip representation. The results are shown in Table \ref{tbl:grulstm}, it can be seen that LSTM outperforms GRU consistently. 

\begin{table}[h!]\small
\centering
\caption{Comparison between LSTM and GRU with using the ``cell outputs average" as the clip level representation and $D_h=512, T=8$ .}
\label{tbl:grulstm}
\begin{tabular}{l|ccccc}
\hline
\multicolumn{1}{c|}{} & \multicolumn{1}{c}{mAP} & \multicolumn{1}{c}{CMC-1} & \multicolumn{1}{c}{CMC-5} & CMC-10 & CMC-20 \\ \hline
LSTM           &     \textbf{72.0} &	\textbf{80.4}	&\textbf{92.7}&	\textbf{94.9}	& \textbf{96.9} \\ \hline
GRU           &    70.5	& 79.7	& 91.5	&93.8	& 95.3  \\ \hline
\end{tabular}
\end{table}

To test RNN performance with different test sequence lengths $T=2,4,8,16$, we fix hidden state size to be $D_h=512$, use LSTM cell and ``cell outputs average" as the clip representation. The results are shown in Table \ref{tbl:rnnlength}, we can see that  $T=4$ gives slightly accurate results than $T=2$ and $T=8$.

\begin{table}[h!]\small
\centering
\caption{Comparison of different sequence length $T=2,4,8,16$ of the RNN model. LSTM cell is used and ``cell outputs average" is used as the clip representation.}
\label{tbl:rnnlength}
\begin{tabular}{l|ccccc}
\hline
\multicolumn{1}{c|}{} & \multicolumn{1}{c}{mAP} & \multicolumn{1}{c}{CMC-1} & \multicolumn{1}{c}{CMC-5} & CMC-10 & CMC-20 \\ \hline
T=2           &     72.4&	81.4	&92.1&	94.9&	96.4\\ \hline
T=4           &    \textbf{73.9}	 & \textbf{81.6}&	\textbf{92.8}	&94.7	& 96.3   \\ \hline
T=8           &    72.0 &	80.4	&92.7&	\textbf{94.9}	& \textbf{96.9}    \\ \hline
T=16           &    60.3	&71.2	&86.5	&90.8	&93.7  \\ \hline
\end{tabular}
\end{table}

\textbf{Comparison with state-of-the-art methods.} We select the best setting in each model (temporal pooling, temporal attention and RNN), and compare their performance; results are shown in Table \ref{tbl:comparison}. For temporal pooling, we select mean pooling and set the sequence length to $T=4$.; for temporal attention, we select ``Softmax" + ``spatial conv + temporal conv" and set $T=4$; for RNN, we set $D_h=512, T=4$ and use ``cell outputs average". We also list the image-based baseline model in Table \ref{tbl:comparison}. We can see that the performance of RNN is even lower than that of image-based baseline model, which implies that using RNN for temporal aggregation is less effective. Temporal attention gives slightly better performance than mean pooling, and outperforms the image-based baseline model by 3\%, which shows the  effectiveness of temporal modeling.
State-of-the-art methods are also listed in Table \ref{tbl:comparison}. It can be seen that our image-based baseline model (mAP=74.1\%) already outperforms state-of-the-art model \cite{hermans2017defense} (mAP=67.7\%) by a large margin; we believe that the improvement mainly comes from the use of Softmax cross-entropy loss. 
We also list the performance after re-ranking \cite{zhong2017re} in Table \ref{tbl:comparison}, which brings another 7.8\% improvement in mAP and 1.7\% in CMC-1.

\textbf{Discussion.} 
Based on the experiments, it can be seen that mean pooling gives 3\% improvement over the image baseline model, which indicates that modeling temporal information on clip level is effective for video-based person reID. Comparing RNN with mean pooling and image baseline model, we can see that RNN's performance is inferior, even worse than the image-based baseline model, which indicates that RNN is either ineffective way to capture temporal information, or too hard to train on this dataset. The reason that previous work \cite{mclaughlin2016recurrent,yan2016person} shows improvement by using RNNs may be that image feature extractor they use are from a shallow CNN, which is an inferior model itself compared to dedicated designed pre-trained CNNs \cite{he2016deep,simonyan2014very}, and RNN performs like an additional layer to extract more representative features. Our temporal attention gives slightly better performance than mean pooling, the reason may be that mean pooling is already good enough to capture information in a clip, as a clip only only contains a few frames, which is 1/4 to 1/2 second, it's difficult to observe any image quality change in such a short period. However, the quality difference among clips can be very large, as the whole video can be very long, thus, a possible future direction is that how to aggregate clip level information (our current solution averages all clips).

\begin{table}[h!]\small
\centering
\caption{Comparison of temporal pooling, temporal attention and RNN with the image-based baseline model. }
\label{tbl:comparison}
\begin{tabular}{l|ccccc}
\hline
\multicolumn{1}{c|}{} & \multicolumn{1}{c}{mAP} & \multicolumn{1}{c}{CMC-1} & \multicolumn{1}{c}{CMC-5} & CMC-10 & CMC-20 \\ \hline
Zheng \etal \cite{zheng2016mars} & 45.6 & 65.0	& 81.1	& - & 88.9 \\\hline
Li \etal \cite{li2017learning} & 56.1 & 71.8 & 86.6 & - & 93.1 \\ \hline
Liu \etal \cite{liu2017quality} & 51.7 & 73.7&	84.9 &-&	91.6	\\\hline
Zhou \etal \cite{zhou2017see} & 50.7 & 70.6 &	90.0 &	-& 97.6	\\\hline
Hermans \etal \cite{hermans2017defense} & 67.7 & 79.8 &	91.4 &	- & - \\\hline

Ours (image)           &      74.1 &	81.3&	92.6&	94.8&	96.7\\ \hline
Ours (3Dconv)           &    70.5 &	78.5	& 90.9	& 93.9	& 95.9   \\ \hline
Ours (pool)          &    76.5	&83.3	&93.0	&95.3	&96.8      \\ \hline
Ours (att)           & \textbf{76.7} &	\textbf{83.3}	& \textbf{93.8}	& \textbf{96.0}	& \textbf{97.4} \\ \hline
Ours (RNN)           &    73.9	 & 81.6&	92.8	&94.7	& 96.3 \\ \hline
\hline
Hermans\etal (re-rank) \cite{hermans2017defense} & 77.4 & 81.2 	&90.7	& - &-	 \\\hline
Ours (re-rank) & \textbf{84.5} & \textbf{85.0} & \textbf{94.7} & \textbf{96.6} & \textbf{97.7} \\\hline

\end{tabular}
\end{table}

\section{Conclusion}
Video-based person reID is an important task, which has received much attention in recent years. We comprehensively study and compare four different temporal modeling methods for video-based person reID, including temporal pooling, temporal attention, RNN and 3D convnets. To directly compare these methods, we fix the base network architecture (ResNet-50) and the loss function (triplet loss and softmax cross-entropy loss). We also propose a new attention generation network which adopts temporal convolutional layers to extract temporal information among frames. The evaluation is done on the MARS dataset. Experimental results show that RNN's performance is inferior, even lower than the image baseline model; temporal pooling can bring 2\%-3\% improvement over the baseline; our temporal-conv-based attention model achieves the best performance among all temporal modeling methods.
\bibliography{egbib}
\end{document}